%% file: root.tex
\documentclass[letterpaper, 10 pt, conference]{ieeeconf}  % Comment this line out if you need a4paper

\IEEEoverridecommandlockouts                              % This command is only needed if 
\usepackage{hyperref}
\hypersetup{
    colorlinks=true,
    linkcolor=black,    
    urlcolor=black,
    }
\usepackage{caption}
\usepackage[font=small]{caption}
\usepackage{cite}
\usepackage{graphicx}
\usepackage{booktabs}
\usepackage{epsfig} % for postscript graphics files
\usepackage{amsmath} % assumes amsmath package installed
\usepackage{color}
\usepackage{dsfont}
\usepackage{bm}
\usepackage{float}
\usepackage{amssymb}
\usepackage{multirow}
\usepackage{etoolbox}
\makeatletter
\patchcmd{\@makecaption}
  {\scshape}
  {}
  {}
  {}
\makeatletter
\patchcmd{\@makecaption}
  {\\}
  {.\ }
  {}
  {}
\makeatother

\usepackage{xspace}
\usepackage{bm}
\usepackage{amsfonts} 
\usepackage{amsmath}

\usepackage{graphicx} 

\usepackage{amsfonts} 
\title{\LARGE \bf
Tac2Motion: Contact-Aware Reinforcement Learning with Tactile Feedback for Robotic Hand Manipulation
}

\author{Yitaek Kim, Casper Hewson Rask and Christoffer Sloth
\thanks{All authors are with The Maersk Mc-Kinney Moller Institute, University of Southern Denmark, Denmark {\tt\small \{yik,chsl\}@mmmi.sdu.dk} \tt\small \{crask21\}@student.sdu.dk.}}

\usepackage{tikz}

\newcommand\submittedtext{%
  \footnotesize This work has been submitted to the IEEE for possible publication. Copyright may be transferred without notice, after which this version may no longer be accessible.}

\newcommand\submittednotice{%
\begin{tikzpicture}[remember picture,overlay]
\node[anchor=south,yshift=10pt] at (current page.south) {\fbox{\parbox{\dimexpr\textwidth-\fboxsep-\fboxrule\relax}{\submittedtext}}};
\end{tikzpicture}%
}

\begin{document}
\maketitle
\submittednotice
\thispagestyle{empty}
\pagestyle{empty}

%%%%%%%%%%%%%%%%%%%%%%%%%%%%%%%%%%%%%%%%%%%%%%%%%%%%%%%%%%%%%%%%%%%%%%%%%%%%%%%%
\begin{abstract}
This paper proposes Tac2Motion, a contact-aware reinforcement learning framework to facilitate the learning of contact-rich in-hand manipulation tasks, such as removing a lid. To this end, we propose tactile sensing-based reward shaping and incorporate the sensing into the observation space through embedding. The designed rewards encourage an agent to ensure firm grasping and smooth finger gaiting at the same time, leading to higher data efficiency and robust performance compared to the baseline. We verify the proposed framework on the opening a lid scenario, showing generalization of the trained policy into a couple of object types and various dynamics such as torsional friction. Lastly, the learned policy is demonstrated on the multi-fingered robot, Shadow Robot, showing that the control policy can be transferred to the real world. The video is available: \url{https://youtu.be/poeJBPR7urQ}.
\end{abstract}

%%%%%%%%%%%%%%%%%%%%%%%%%%%%%%%%%%%%%%%%%%%%%%%%%%%%%%%%%%%%%%%%%%%%%%%%%%%%%%%%
\input{introduction}

\input{related_works}
\input{proposed_method}
\input{simulation}
\input{experimental_setup}
%\input{results}
%l\input{discussion}
\input{conclusions}
%%%%%%%%%%%%%%%%%%%%%%%%%%%%%%%%%%%%%%%%%%%%%%%%%%%%%%%%%%%%%%%%%%%%%%%%%%%%%

\newpage
\renewcommand{\thefigure}{A.\arabic{figure}}
\setcounter{figure}{0}

\section*{APPENDIX}
This section includes the detailed information about object geometries and simulation parameters. 

\subsection{Details in Simulation}
In the simulation, we use several object geometries such as cylinder, square, and hexagon with contact guide fixtures in the following:
\begin{figure}[h]
    \centering
    \includegraphics[width=1\linewidth]{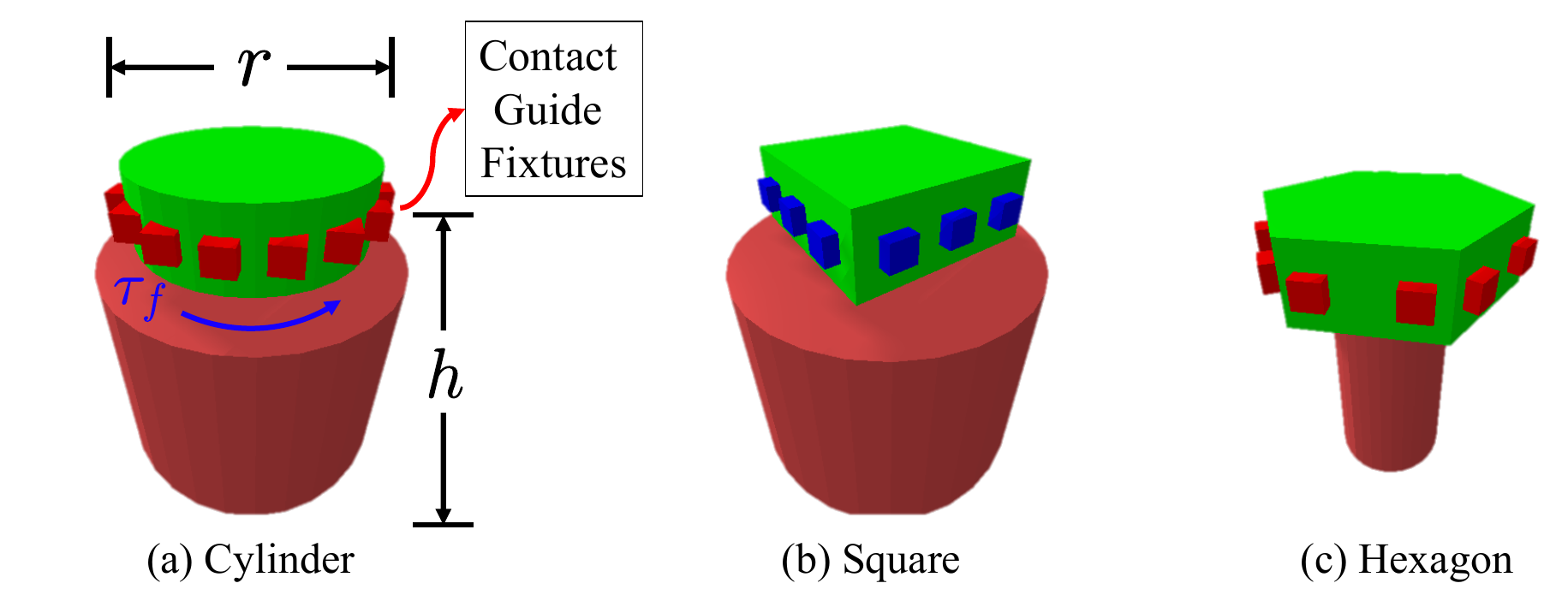}
    \caption{Types of the lid in the simulation. All lids have multiple virtual boxes on their rims as the contact guides.}
    %\vspace{-0.3cm}
    \label{sim:types}
\end{figure}

We apply minimal domain randomization to evaluate the proposed framework. Domain randomization includes action noise, joint measurement noise, and torsional friction of a lid. The detail ranges of each component are presented in the following: 

\begin{table}[h]
\centering
\begin{tabular}{lc}
\hline\hline
Action Noise $\quad \quad \quad$ & $~\mathcal{N}(0,0.2)$ \\ 
Joint Measurement Noise $\quad \quad \quad$ & $~\mathcal{N}(0,0.4)$ \\ 
Torsional Friction $\quad \quad \quad$ & [0.9,1.5] \\ \hline\hline
\end{tabular}
\caption{Domain Randomization Parameters}
\end{table}

The parameters used in the simulation are shown in the following table: 
\begin{table}[h]
\centering
\begin{tabular}{lc|lc}
\hline
\hline
Maximum episode length $\quad \quad$ & 1000   & $\epsilon$$\quad \quad \quad$     & 0.005 \\
Batch size             & 10240  & $\rho$          & 0.06  \\
Number of environments & 10240  & $\lambda_{\textnormal{cpr}}$ & 8.0   \\
Clip observation       & 10.0   & $\lambda_{\textnormal{crr}}$ & 2.0   \\
Clip action            & 1.0    & $\lambda_{\textnormal{rr}}$ & 850.0 \\
Torsional stiffness    & 0.5    & $\lambda_{\textnormal{angle}}$ & 20.0  \\
Torsional damping      & 3.0    & $\lambda_{\textnormal{action}}$ & 0.001 \\
Contact offset         & 0.002  & $\lambda_{\textnormal{work}}$ & 1.0   \\
dt                     & 0.0166 & $\lambda_{\textnormal{gaiting}}$ & 8.0   \\
$\eta$                 & 0.75   & Action scale   & 0.1   \\ \hline\hline
\end{tabular}
\caption{Simulation Parameters}
\end{table}

%%%%%%%%%%%%%%%%%%%%%%%%%%%%%%%%%%%%%%%%%%%%%%%%%%%%%%%%%%%%%%%%%%%%%%%%%%%%%%%%

\bibliographystyle{IEEEtran}
\bibliography{bibliography}
\end{document}

%% file: introduction.tex
\section{Introduction}\label{sec:introduction}
Robotic hand manipulation often requires contact-rich dexterity, which has long remained an exceptional challenge due to complex high dimensionality, lack of sensing modality, and limited generalization. Rapid progress in Reinforcement Learning (RL) has accelerated a major step to overcome the above issues, which in turn has enhanced robotic in-hand manipulation capabilities \cite{yin2025learninginhandtranslationusing}. In addition, visual feedback often assists in-hand manipulation control policies, showing outstanding performances \cite{SridharICRA2023,HoloSridharICRA2023}. 

Even though using visual sensory feedback into RL is widely used, it is still limited by occlusion from fingers. To overcome this, one promising approach is to add extra cameras \cite{li2025dropdexterousreorientationonline}, which is unrealistic in practice. Unlike the vision system, tactile sensory feedback is free from the occlusion issue and is a more natural way to obtain useful contact information efficiently. Inspired by this, human demonstrations with the tactile sensor technology (e.g. GelSight \cite{YuanGelsight2017}) have enabled learning human's dexterity efficiently \cite{iyer2024open, HuangTMech2025}. However, employing human data is quite expensive due to the kinematic and dynamic embedding problem \cite{pmlr-v164-chen22a}. 

% Moreover, collecting pure tactile information from human operators, not the sensors of robotic hands has remained a persistent obstacle \cite{HandaICRA2020,LuoICRA2024}. 

%Not surprisingly, it is well aligned with how human uses tactile sensation for fine motor skills in daily life \cite{Johansson2009}.

To get around these barriers, model-free RL has provided a viable alternative for learning contact-rich dexterity, showing meaningful performance in multi-finger manipulation domain with tactile sensing  \cite{ChenTransactions2024, OpenAI2019, openai2019rubiks, ZhuICRA2019, YangRAL2023, GuzeyICRA2024}. Despite the remarkable progress in RL-based dexterous frameworks with tactile sensing, many of them have often neglected the benefits from using tactile sensors in shaping reward functions for learning a dexterous control policy, and the related research has not yet been fully investigated.

% Instead of using vision, tactile sensory feedback also can be incorporated in RL-based approaches as observations, improving the task completion performances \cite{YangRAL2023, GuzeyICRA2024}. 

\begin{figure}[t]
    \centering
    \includegraphics[width=1\linewidth]{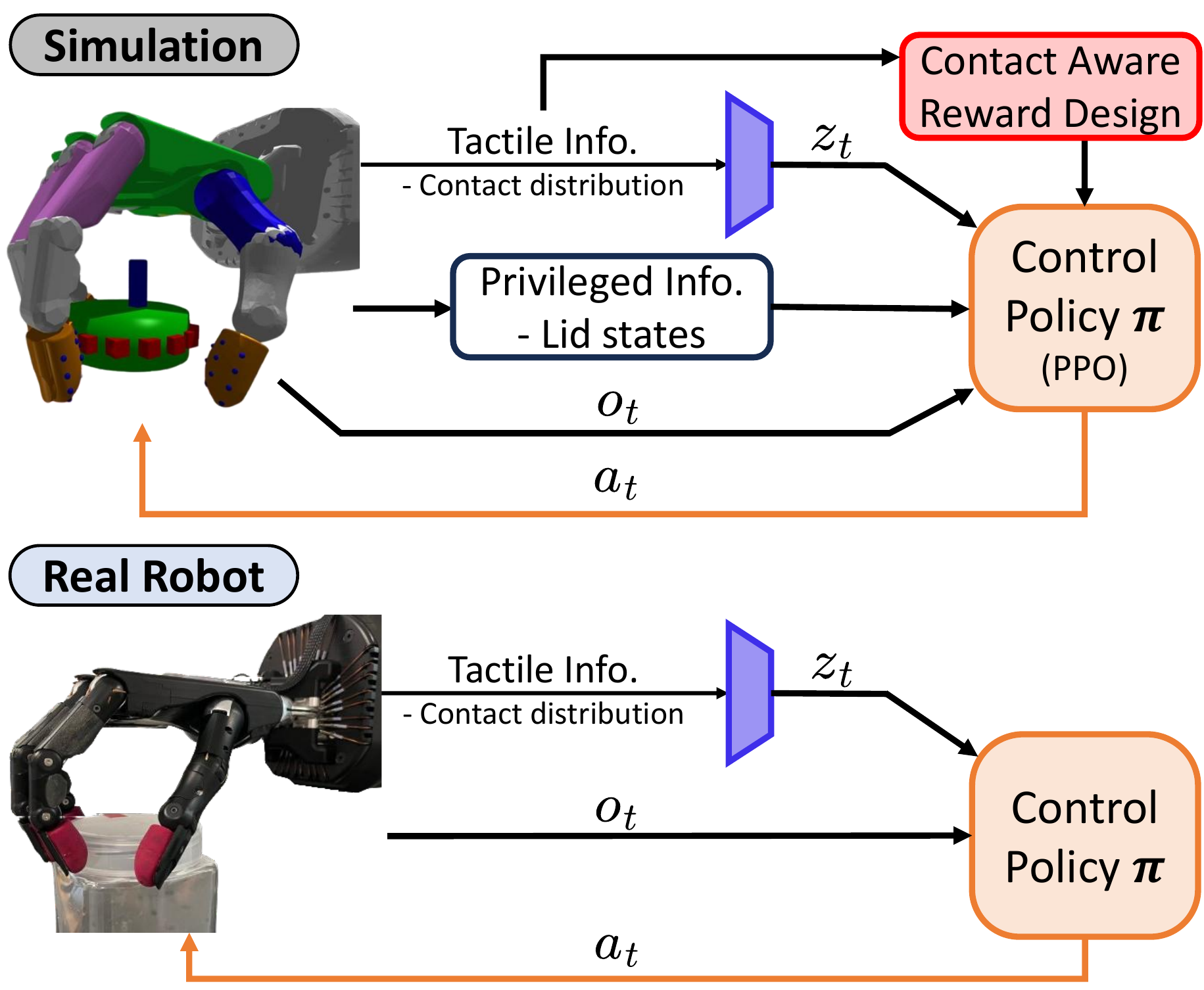}
    \caption{Outline of Tac2Motion including model training and transferring for the contact aware control policy.}
    \vspace{-0.3cm}
    \label{pro:overview}
\end{figure}

 In this work, we propose a new RL-based framework for in-hand manipulation in contact-rich environments such as lid-removing tasks. To this end, the main idea is to incorporate tactile sensing into observation and reward functions to guide contact-aware control policy. We approximate and simulate tactile sensing through multiple contact penetrations. Lastly, the proposed framework is demonstrated on removing a lid of the bottle, including a comparison to the baseline. The contributions are outlined as follows: 

\subsection{Contributions}
 \begin{itemize}
     \item We utilize tactile sensing in the observation space to estimate the object dynamics such as torsional friction of a lid. This allows an agent to adapt to uncertain dynamics of the lid.
     \item We leverage the contact pressure captured by tactile sensing to shape the reward function that maximizes friction between the fingers and the lid. This shows a major departure from the existing use of tactile sensing modality in RL.
     \item We propose a computationally efficient technique to emulate the effects caused by soft finger contact interaction at the patch level. To this end, we introduce a virtual torque on the lid induced by contact friction, which is generated from contact pressure. This eliminates the necessity of complex patch contact modeling and enables efficient and stable training of removing the lid based on tactile-inspired dynamics.
 \end{itemize}

%% file: proposed_method.tex
\section{Contact-Aware Reinforcement Learning \\ For Dexterous Hand Manipulation}\label{sec:method}
We aim to learn contact-aware torsional motion for opening a lid through RL since combining tactile perception and dynamic dexterity in a single control pipeline is challenging. During the task, it is necessary to balance between finger gaiting and maintaining sufficient patch contact to prevent slippage. This problem is formulated as Markov Decision Process under partially observable states.

\subsection{Learning Control Policy}
During training the control policy, agents observe $\bm{o}_t$ including joint position, $\bm{q}_t$, joint angular velocity, $\dot{\bm{q}}_t$, previous joint target position, $\bm{q}^d_{t-1}$, tactile embedding, $\bm{z}_t$, and the center position of a lid, $\bm{p}$ at each time step, $t$ as shown in Fig.~\ref{pro:overview}. Action space includes a relative desired joint position, $\bm{a}_t$, and the final target joint position is decided by adding ${\bm{a}_t}$ to $\bm{q}^d_{t-1}$. To avoid jerky motions, we use Exponential Moving Average (EMA) function to make the target values smooth, $\bm{q}^d_{t} = \bm{q}^d_{t-1} + \eta\textnormal{EMA}(\bm{a}_t)$ where $\eta$ is a scaling parameter \cite{pmlr-v270-lin25c}. Lastly, we utilize PPO to train the control policy, accompanying asymmetric learning and actor-critic structure.

\subsection{Reward Design}
\begin{figure}[t]
    \centering
    \includegraphics[width=1\linewidth]{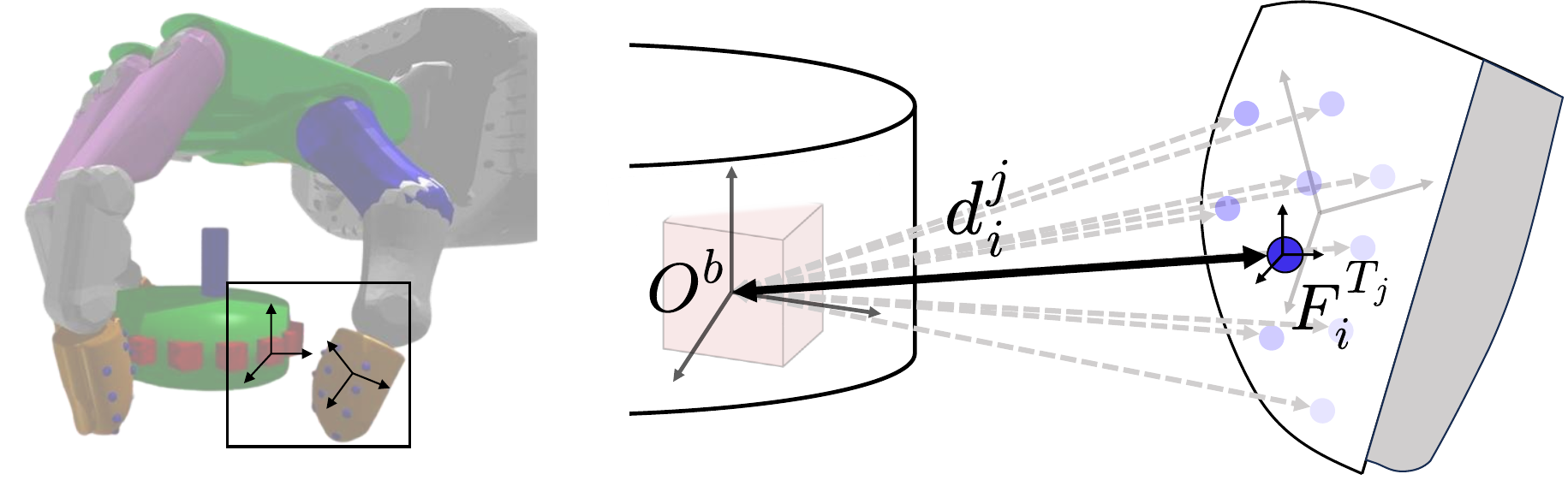}
    \caption{Illustration of distance, $d^j_i$ between contact reference object base and tactile sensors. The contact reference object base are represented by the red boxes and the blue dots are tactile sensors.}
    \vspace{-0.3cm}
    \label{pro:contact_d}
\end{figure}

We design the following contact-aware rewards to induce the torsional hand motion. The individual reward functions are associated with contact pressure, contact release, rotation. 

\textit{1) \textbf{Contact Pressure Reward, (CPR)}}: As the direct simulation of contact pressure is non-trivial, we formulate the reward function based on contact penetration. Let $O^b \in \mathbb{R}^{m\times 3}$ be the contact reference object frames attached to the lid as shown in Fig.~\ref{pro:contact_d}. Let $F^{T}_i \in \mathbb{R}^{k \times 3}$ be the points of $k$ tactile sensors on each finger, $i$. For an each sensor, $F^{T_j}_i, \quad j=\{1,\dots,k\}$, the closest distance from contact reference frames is obtained based on Euclidean distance, $d^j_i(O^b,F^{T_j}_i)$ where $d^j_i(A,B^j) = \min_{n} {||A^n-B^j||}$ as shown in Fig.~\ref{pro:contact_d}. 

To detect contact states, we set the contact threshold, $\epsilon\in \mathbb{R}_{>0}$, and then define contact penetration distance, $cd^j_i$ for $\forall i,j$ as:
\begin{equation}
    \quad cd^j_i =\begin{cases}
			d^j_i(O^b,F^{T_j}_i), & d^j_i \leq \epsilon \\
            0, & \text{otherwise}.
		 \end{cases} 
         \nonumber
\end{equation}

Subsequently, to increase learning stability, we normalize each contact penetration to the range $[0, 1]$ with respect to the maximum penetration on each finger. Consequently, we derive the contact pressure reward as
\begin{equation}
     r_{\textnormal{cpr}} = \sum_{i}^{5} \sum_{j}^{k} cd^j_i \triangleq \sum_{i}^{5} G_i,
\end{equation}
while introducing \textit{grasping quality}, $G_i \in \mathbb{R}$ which indicates how well the contact is established at each finger, $i$. $G_i$ will be further used to maximize the rotation reward. 

\textit{2) \textbf{Contact Release Reward, (CRR)}}: It is crucial to release and regrasp the fingers to generate continuous rotation movement of the lid, thereby, we define the following contact release reward on each finger as: 
\begin{equation}
    r_{\textnormal{crr}}  = \sum_{i}^{5} c^i,
\end{equation}
where $c^i$ is the binary variable to detect the contact-release transition, determined from:
\begin{equation}
    \quad c^i =\begin{cases}
			1, & \text{if contact happens in the previous step}\\
            0, & \text{otherwise}.
		 \end{cases} 
         \nonumber
\end{equation}

\textit{\textbf{3) Rotation Reward, (RR)}}: We newly define the torsional reward designed to promote grasping and rotation at the same time, extending the previous twisting reward proposed \cite{pmlr-v270-lin25c}:
\begin{equation}
    r_{\textnormal{rr}} = \sum_{i}^5 G_i\Delta q_{\textnormal{lid}},
\end{equation}
where $\Delta q_{\textnormal{lid}} = q_{\textnormal{lid}}^{t} - q_{\textnormal{lid}}^{t-1}$ is the rotation difference between the rotation angles of the lid during one step transition. The reward is determined with the hyperparameters to balance each objective and stabilize learning: 
\begin{equation}
    r \doteq \lambda_{\textnormal{cpr}}r_{\textnormal{cpr}} + \lambda_{\textnormal{crr}}r_{\textnormal{crr}} + \lambda_{\textnormal{rr}}r_{\textnormal{rr}},\label{final_r}
\end{equation}
where $\lambda_{\textnormal{cpr}}, \lambda_{\textnormal{crr}}, \lambda_{\textnormal{rr}} \in \mathbb{R}_{> 0}$.

\subsection{Penalty Design}
Adding penalty terms helps to discourage the undesired behaviors such as incorrect rotation direction and jerky finger motions. We also consider the auxiliary penalty terms used in \cite{pmlr-v270-lin25c} in the following: 

\textit{\textbf{1) Angle Direction Penalty}}: The axis of the lid should be aligned with the axis of the rotation movement of the lid from the fingers. The angle direction penalty term is defined as $r_{\textnormal{angle}} = -\arccos(\langle \bm{z}_{\textnormal{lid}}, \bm{z}\rangle )$ where $\bm{z}_{\textnormal{lid}}, \bm{z} \in \mathbb{R}^3$ are the axis of the lid and the desired rotation direction, respectively.

\textit{\textbf{2) Action Penalty}}: To get around abrupt actions, we consider the action penalty which is defined as:
\begin{equation}
    r_{\textnormal{action}} = - ||\bm{a}_t||^2,
\end{equation}
where $\bm{a}_t \in \mathbb{R}^{22}$ is the relative desired joint position.

\textit{\textbf{3) Work Efficiency Penalty}}: We prevent the agent from applying the unnecessary excessive control efforts. The term is defined as: 
\begin{equation}
    r_{\textnormal{work}} = - \sum|\bm{\tau}_t^{\top}\Delta \bm{q}_t|,
\end{equation}
where $\bm{\tau}_t \in \mathbb{R}^{22}$ is the control torque and $\Delta\bm{q}_t\in \mathbb{R}^{22}$ is the change of joint positions.

\textit{\textbf{4) Gaiting Penalty}}: We introduce gaiting penalty to guide efficient torsional motions in the proposed framework. Forces applied by fingers should be able to produce a positive torque toward the opening direction of a lid. And the movements of each finger should not be interrupted each other. The term is defined as:

\begin{equation}
    r_{\textnormal{gaiting}} =  \sum_{i}^{5} \textnormal{sign}(w^z_i)G_i,
\end{equation}
where $w^z_i$ is the $z$-axis component of $\bm{w}_i \in \mathbb{R}^3$ which is the cross product between $\bm{v}_i \in \mathbb{R}^3$ and $\bm{p}^{\textnormal{center}}_i \in \mathbb{R}^3$, $\bm{w}_i = \bm{v}_i \times \bm{p}^{\textnormal{center}}_i$. $\bm{v}_i$ is the velocity of the tip of each finger, and $\bm{p}^{\textnormal{center}}_i$ is the vector from the center of the lid and each tip. Since the base frame is with respect to the center of the lid, only z-component of $\bm{w}_i$ is used to decide the resulting torque directions by each finger. Since the stronger normal forces enhance the grasping quality, $G_i$ and generates higher contact frictions and in turn  the larger torques of the lid, we multiply $G_i$ in the penalty term. This term ensures that fingers can move simultaneously in the opening direction of the lid. Therefore, the final reward function is defined as:
\begin{equation}
    r_{\textnormal{{final}}} \doteq r + \lambda_{\textnormal{angle}}r_{\textnormal{angle}} + \lambda_{\textnormal{action}}r_{\textnormal{action}} + \lambda_{\textnormal{work}}r_{\textnormal{work}} +\lambda_{\textnormal{gaiting}}r_{\textnormal{gaiting}},\label{f_r}
\end{equation}
where $\lambda_{\textnormal{angle}}, \lambda_{\textnormal{action}}, \lambda_{\textnormal{work}}, \lambda_{\textnormal{gaiting}} \in \mathbb{R}_{>0}.$

%\begin{itemize}
    % \item \yk{Contact Release Reward}: Give a reward when the fingers contacted and then were released. This is to generate next motion. Otherwise, the fingers continuously contacted to the bottle and the rotation does not happen.
    
    %\item \yk{Rotation Reward}: Give a reward when the rotation occurs. This is the highest reward.
    %\item  \yk{Multi-Finger Force Distribution Reward}: Reward if pressure is evenly distributed across multiple fingers.
    %\item \yk{Avoiding finger's collision}: To avoid collisions from each finger, provide a reward to make the fingers atay far apart each other. 
    % \item \yk{Slippage Penalty}:
    % Apply a penalty if there is excessive movement in the contact area over time (e.g., high shear force or displacement).
%\end{itemize}

% \begin{itemize}
%     \item \yk{Angle Direction Penalty}: 
%     \item \yk{Work Efficiency Penalty}: 
%     \item \yk{Action Penalty}:
%     %\item  \yk{Multi-Finger Force Distribution Reward}: Reward if pressure is evenly distributed across multiple fingers.
%     %\item \yk{Avoiding finger's collision}: To avoid collisions from each finger, provide a reward to make the fingers atay far apart each other. 
%     % \item \yk{Slippage Penalty}:
%     % Apply a penalty if there is excessive movement in the contact area over time (e.g., high shear force or displacement).
% \end{itemize}

\subsection{Termination Condition Design}
To facilitate more efficient learning, the episode is immediately terminated and reset when the fingertips move too far away from the lid. We reset the episode when $\max_{i} \big(\max_{j}d^j_i \big) \geq \rho$ for $\forall i,j$, where $\rho \in \mathbb{R}_{>0}$ is the termination threshold. The detail domain randomization is presented on Appendix.
% \begin{itemize}
%     \item \yk{non-contact fingertip behavior}: terminate the episode when the fingertips is moving far way from the object. (threshold-based)
% \end{itemize}

% \subsection{Tactile Adaption Design}

% \subsection{Domain Randomization}

% \subsection{\yk{possible contribution:} KAN-PPO}
% \cite{10773080} We should try KAN and MLP both. And we can provide comparison of performance with respect to calculation time and the number of parameters. 

%% file: simulation.tex
\section{Simulation}\label{sec:simulation}
In this section, we carry out several simulations to investigate the performance of the learned contact-aware in-hand manipulation in the lid-removal scenario. We evaluate the proposed method based on the following metrics: Rotation Score (RS), Rotation Time (RT), and Success Rate (SR). We then compare the proposed method with the baseline provided by \cite{pmlr-v270-lin25c} and conduct an ablation study to evaluate the combination of reward functions.

\begin{figure*}[t]
    \centering
    \includegraphics[width=1\linewidth]{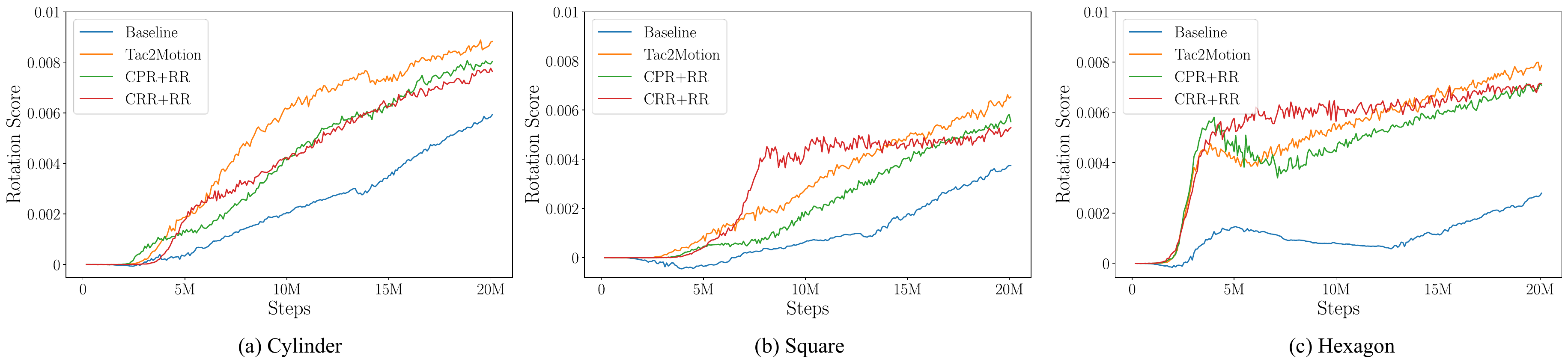}
    %\vspace{-0.7cm}
    \caption{Training performance of each method over 20M steps across different types, sizes, and object dynamics. The results show that Tac2Motion learns the torsional motion faster then other methods and achieves superior final performance.}
    %\vspace{-0.2cm}
    \label{sim_graphs}
\end{figure*}

\begin{table*}[t]
\centering
\begin{tabular}{c|ccc@{\hskip 5pt}c|ccc@{\hskip 5pt}c|ccc@{\hskip 5pt}c}
\hline
\hline
\multirow{2}{*}{Methods}    & \multicolumn{4}{c|}{Rotation Score (RS)}    & \multicolumn{4}{c|}{Rotation Time (RT) [s/rev]}       & \multicolumn{4}{c}{Success Rate (SR)}        \\ \cline{2-13} 
                            & Cylinder & Square & Hexagon & Average                & Cylinder & Square & Hexagon & Average                & Cylinder & Square & Hexagon & Average                \\ \hline
\multirow{2}{*}{Baseline \cite{pmlr-v270-lin25c}}   & $0.0087$  & $0.005$  & $0.0035$  & \multirow{2}{*}{$0.0057$} & $3.88$  & $6.29$  & $5.68$  & \multirow{2}{*}{$5.28$} & \multirow{2}{*}{$0.38$}  & \multirow{2}{*}{$0.35$}  & \multirow{2}{*}{$0.05$}  & \multirow{2}{*}{$0.26$} \\
                            & \scriptsize{$\pm0.0007$}  & \scriptsize{$\pm0.0005$}   & \scriptsize{$\pm 0.002$}  &                        & \scriptsize{$\pm3.26$}   & \scriptsize{$\pm3.45$}   & \scriptsize{$\pm2.46$}   &                          \\ \hline
\multirow{2}{*}{Tac2Motion} & $\bm{0.0108}$  & $\bm{0.0086}$  & $\bm{0.0094}$  & \multirow{2}{*}{$\bm{0.0096}$} & $\bm{2.36}$  & $\bm{5.26}$  & $\bm{3.28}$  & \multirow{2}{*}{$\bm{3.63}$} & \multirow{2}{*}{$\bm{0.78}$}  & \multirow{2}{*}{$\bm{0.72}$}  & \multirow{2}{*}{$\bm{0.73}$}  & \multirow{2}{*}{$\bm{0.74}$} \\
                            & \scriptsize{$\pm0.0004$}  & \scriptsize{$\pm0.0007$}   & \scriptsize{$\pm0.0007$}  &                        & \scriptsize{$\pm1.30$}   & \scriptsize{$\pm2.48$}   & \scriptsize{$\pm1.11$}   &                        \\ \hline
\multirow{2}{*}{CPR+RR}     & $0.0097$  & $0.0078$  & $0.008$  & \multirow{2}{*}{$0.0085$} & $3.24$  & $6.17$  & $3.83$  & \multirow{2}{*}{$4.41$} & \multirow{2}{*}{$0.48$}  & \multirow{2}{*}{$0.54$}  & \multirow{2}{*}{$0.53$}  & \multirow{2}{*}{$0.51$} \\
                            & \scriptsize{$\pm0.0003$}  & \scriptsize{$\pm0.0005$}   & \scriptsize{$\pm0.0009$}  &                        & \scriptsize{$\pm1.97$}   & \scriptsize{$\pm3.17$}   & \scriptsize{$\pm1.45$}   &                        \\ \hline
\multirow{2}{*}{CRR+RR}     & $0.009$  & $0.0068$  & $0.007$  & \multirow{2}{*}{$0.0076$} & $3.08$  & $6.5$  & $5.71$  & \multirow{2}{*}{$5.09$} & \multirow{2}{*}{$0.49$}  & \multirow{2}{*}{$0.39$}  & \multirow{2}{*}{$0.10$} & \multirow{2}{*}{$0.32$} \\
                           & \scriptsize{$\pm0.0003$}  & \scriptsize{$\pm0.0007$}   & \scriptsize{$\pm0.0005$}  &                        & \scriptsize{$\pm2.05$}   & \scriptsize{$\pm3.84$}   & \scriptsize{$\pm2.95$}   &                          \\ \hline\hline

\end{tabular}
\caption{Simulation results. Rotation Score (RS) is the average of the difference between the previous and current angles, and Rotation Time (RT) shows the time taken for one revolution, which is an indicator of how fast the lid is rotating. Success Rate (SR) indicates that the lid is successfully rotated for one round within the specified time.}
\label{simulation_results}
%\vspace{-0.2cm}
\end{table*}

\begin{figure*}[t]
    \centering
    \includegraphics[width=1\linewidth]{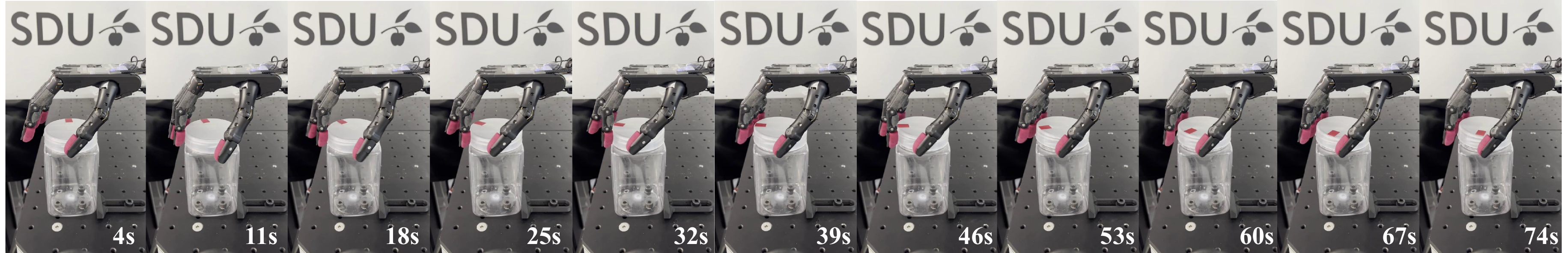}
    %\vspace{-0.7cm}
    \caption{Snapshots from experimental deployment of the learned control policy for removing the lid with a multi-fingered robotic hand.}
    \vspace{-0.4cm}
    \label{real_robot}
\end{figure*}

\subsection{Simulation Setups}
We conduct the simulation in IsaacGym by modifying the code in \cite{pmlr-v270-lin25c}. We use multi-fingered robotic hand, Shadow Robot to assess the performance of each method and also leverage the guided contact fixtures suggested in \cite{pmlr-v270-lin25c} on the lids as shown in Fig.~\ref{sim:types}. Different geometries (e.g. cylinder, square, and hexagon), sizes, frictions of lids are considered in our simulations. We train each method with $10240$ environments and batch sizes, $10240$ with one single NVIDIA GeForce RTX 4090 GPU, which takes $20$ minutes for the entire training. Afterwards, we test the learned models on $100$ agents to show general performances. Note that the tactile embedding is frozen on training and testing.

\subsection{Virtual Torque}
Simulating patch contact interactions and their effects (e.g. increase of contact friction) is non-trivial and highly computational. To resolve this, we introduce a virtual torque on a lid that corresponds to the variation of the contact friction caused by the patch contact interaction. The higher the normal magnitude of each tactile sensor, the more it is likely to increase the friction, which in turn generates a stronger torque applied to the lid. This technique is distinguished from the penetration-based friction calculation that is commonly used in the simulation, enabling the practical implementation of soft finger model \cite{Prattichizzo2016}.

\subsection{Simulation Results}
We conduct several simulations to verify Tac2Motion, including an ablation study and comparison to the baseline. First of all, the proposed method enables faster learning than the baseline at approximately twice the speed across all types of lids, showing higher data efficiency as shown in Fig.~\ref{sim_graphs}. This is because Tac2Motion encourages the firm grasping directly, which in turn increases the contact friction and leads to faster rotation, compared to the fingertip–lid distance-based reward that the baseline used. For the ablation study, we have two cases: using either contact pressure (CPR) or contact release (CRR) rewards, each combined with the rotation reward. Both cases show similar performances on the final rotation score, but using CRR exhibits faster convergence to the final rotation score. On the other hand, Tac2Motion shows the superior performance at the final rotation score as shown in Fig.~\ref{sim_graphs}. 

\subsection{Generalization of Tac2Motion}
We compare Tac2Motion with several methods to verify its efficacy, including baseline \cite{pmlr-v270-lin25c}, a couple of combinations of reward functions. We simulate each method on three types of lids, such as cylinder, square, and hexagon, to observe the generalization performance. The multiple metrics to evaluate the performance are leveraged: Rotation Score (RS), Rotation Time (RT), and Success Rate (SR). RS is to analyze the angle changes of the lid per each step, and RT is the average time for one revolution of the lid. We also define success and failure based on whether the lid completes one full revolution within certain limits, $2.5$ [s], $5.0$ [s], and $3.5$ [s] corresponding to cylinder, square, and hexagon, respectively.  As summarized in Table.~\ref{simulation_results}, Tac2Motion shows the superior performances across all metrics. RS and RT indicate that Tac2Motion achieves the fastest rotation of the lid, and SR shows the most robust rotational performance compared to the baseline. These advantages arise because Tac2Motion encourages the agent to behave with firm grasping and smooth finger gaiting simultaneously during rotation, while the baseline primarily focuses on rotation performance. In an ablation study, it is observed that applying CPR reward (CPR+RR) is highly beneficial for removing the lid since maintaining grasping of the lid during rotating ensures stable finger behavior. In contrast, using only CRR (CRR+RR) does not lead to meaningful final performance according to the success rate and rotation score as presented in Table.~\ref{simulation_results}. Therefore, the results of the simulation support that contact-aware reward design improves the performance of contact-rich in-hand manipulation.

% \begin{itemize}
%     \item Twisting lids off with different 20 sizes and frictions 
%     \item Baseline: original reward design, my reward design, my reward design with adaptation layer. 
%     \item Metrics: rotation score,
% \end{itemize}
%\vspace{0.25cm}
% \begin{figure*}[t]
%     \centering
%     \includegraphics[width=1\linewidth]{figure/two_col_sim_results.pdf}
%     \vspace{-0.0cm}
%     \caption{Simulation results  \yk{TBD}}
%     \vspace{-0.0cm}
%     \label{sim:graph}
% \end{figure*}

%% file: experimental_setup.tex
\section{Experiments}\label{sec:experimental_setup}
We utilize a 24-DoF Shadow Robot to test Tac2Motion. The hand is mounted to a UR10e robot arm to facilitate our experiment, and the initial pose of the lid is known in advance. The lid used in the experiment is cylindrical with a radius of $40$ mm. We update and send the joint commands from the learned control policy at $10$ Hz, to the high-level joint control loop that runs at $60$ Hz. To mitigate jerky finger motions, we apply an exponential moving average to the target joint positions before sending them to the robot, which leads to smoother movements in the real robot. The result shows that the learned policy, Tac2Motion can be successfully deployed without requiring any manual tuning in a real robotic setup and smoothly open the cylindrical lid by using all five fingers as shown in Fig.~\ref{real_robot}. In addition, Tac2Motion achieves firm grasping and fast rotation simultaneously, leading to a stable and seamless removal of the lid. 

% For the practical implementation, it is worth noting that gravity compensation should be considered in both simulation and reality. Otherwise, the control policy tends to learn gravity compensation during training, which causes a dynamics mismatch when transferring to the robot.

% "As we mentioned before, the core of Bi-DexHands is to build
% up a learning framework for two Shadow Hands capable of
% diverse skills as humans, such as reaching, throwing, catching,
% picking and placing. To be specific, Bi-DexHands consists of
% three components: datasets, tasks, and learning algorithms, as
% shown in Fig. 3. Varying worlds provide a large number of
% basic settings for robots, including the configuration of robotic
% hands and objects. Meanwhile, a variety of tasks corresponding
% to children’s behaviors at different ages make it possible to learn
% dexterity like a human. Combining a dataset and task, we can
% generate a specific environment or scenario for the following
% learning. Eventually, our experiments demonstrate that reinforcement learning is able to facilitate the robots to achieve some
% remarkable performance on such challenging tasks, and there is
% still some room for improvement and more difficult tasks for
% future work."

% \begin{itemize}
%     \item Twisting lids off with different 4 sizes and friction
%     \item Tightening bolts with different 4 diameters
%     \item Assemble "Danfoss" parts with different 4 parts, but same parts 
%     \item Assemble 
% electric bulb with different 4 sizes
% \end{itemize}

%% file: conclusions.tex
\section{Conclusion}\label{sec:conclusions}
This paper presents Tac2Motion, an RL-based contact-aware learning framework for contact-rich in-hand manipulation. Within the framework, we introduce a new contact-aware reward and observation designs based on tactile perception, which in turn accomplish firm grasping and smooth finger gaiting simultaneously. We verify that the learned control policy outperforms the baseline in opening a lid, observing higher data efficiency and robust behaviors. Lastly, we demonstrate the learned control policy on a real robotic setup and show that it can actually remove the lid smoothly. For future work, we plan to further investigate tactile adaptation to various dynamics of the lid and aim to carry out a wide range of experiments across various real objects under different conditions. 

\section{Acknowledgement}
The work was supported by Fabrikant Vilhelm Pedersen og Hustrus Legat.